\documentclass[conference]{IEEEtran}
\IEEEoverridecommandlockouts

\usepackage{multirow}

\usepackage{bm}
\usepackage{enumitem}

\usepackage{amsmath}
\usepackage{amssymb}
\usepackage{mathtools}
\usepackage{caption}

\usepackage{subcaption}
\usepackage{subfiles}

\usepackage{cite}
\usepackage{amsmath,amssymb,amsfonts}
\usepackage{algorithm}
\usepackage{algorithmic}
\usepackage{graphicx}
\usepackage{textcomp}
\usepackage{xcolor}

\usepackage{ulem}
\def\BibTeX{{\rm B\kern-.05em{\sc i\kern-.025em b}\kern-.08em
    T\kern-.1667em\lower.7ex\hbox{E}\kern-.125emX}}
\begin{document}

\title{Symbol-Temporal Consistency Self-supervised Learning for Robust Time Series Classification }
\author{\IEEEauthorblockN{Kevin Garcia, Cassandra Garza, Brooklyn Berry, Yifeng Gao}
\IEEEauthorblockA{\textit{Department of Computer Science, University of Texas Rio Grande Valley}}
}
\maketitle
\begin{abstract} 

The surge in the significance of time series in digital health domains necessitates advanced methodologies for extracting meaningful patterns and representations. Self-supervised contrastive learning has emerged as a promising approach for learning directly from raw data. However, time series data in digital health is known to be highly noisy, inherently involves concept drifting, and poses a challenge for training a generalizable deep learning model. In this paper, we specifically focus on data distribution shift caused by different human behaviors and propose a self-supervised learning framework that is aware of the bag-of-symbol representation. The bag-of-symbol representation is known for its insensitivity to data warping, location shifts, and noise existed in time series data, making it potentially pivotal in guiding deep learning to acquire a representation resistant to such data shifting. We demonstrate that the proposed method can achieve significantly better performance where significant data shifting exists.
\end{abstract}
\vspace{-3mm}
\section{Introduction}
\vspace{-2mm}

The growing importance of time series data in the digital health domains \cite{kiernan2018accelerometer, howell2015monitoring, liu2021deep}, especially data collected from motion sensors, has fueled the need for advanced methodologies that can extract meaningful patterns. While many representations have been explored \cite{Li2022symbolicrep,yin2015symbolicrep}, Self-Supervised Contrastive Learning(SSL) has emerged as a promising approach \cite{ijcai2021-324, zhang2022selfsupervised, kiyasseh2021selfsupervised} because it trains models without human annotation, and is versatile enough to be applied to tasks such as classification and anomaly detection. However, challenges lie in systematic data shifting that is prevalent in time series data mining \cite{Sun2021neural, Fawaz2019deeplearn}. For example, in sensor readings from human activities, a male may exhibit different behavior compared to a female due to varying weights, paces, and fitness levels. 
 
\begin{figure}[t]
    \centering
   \includegraphics[scale=0.07]{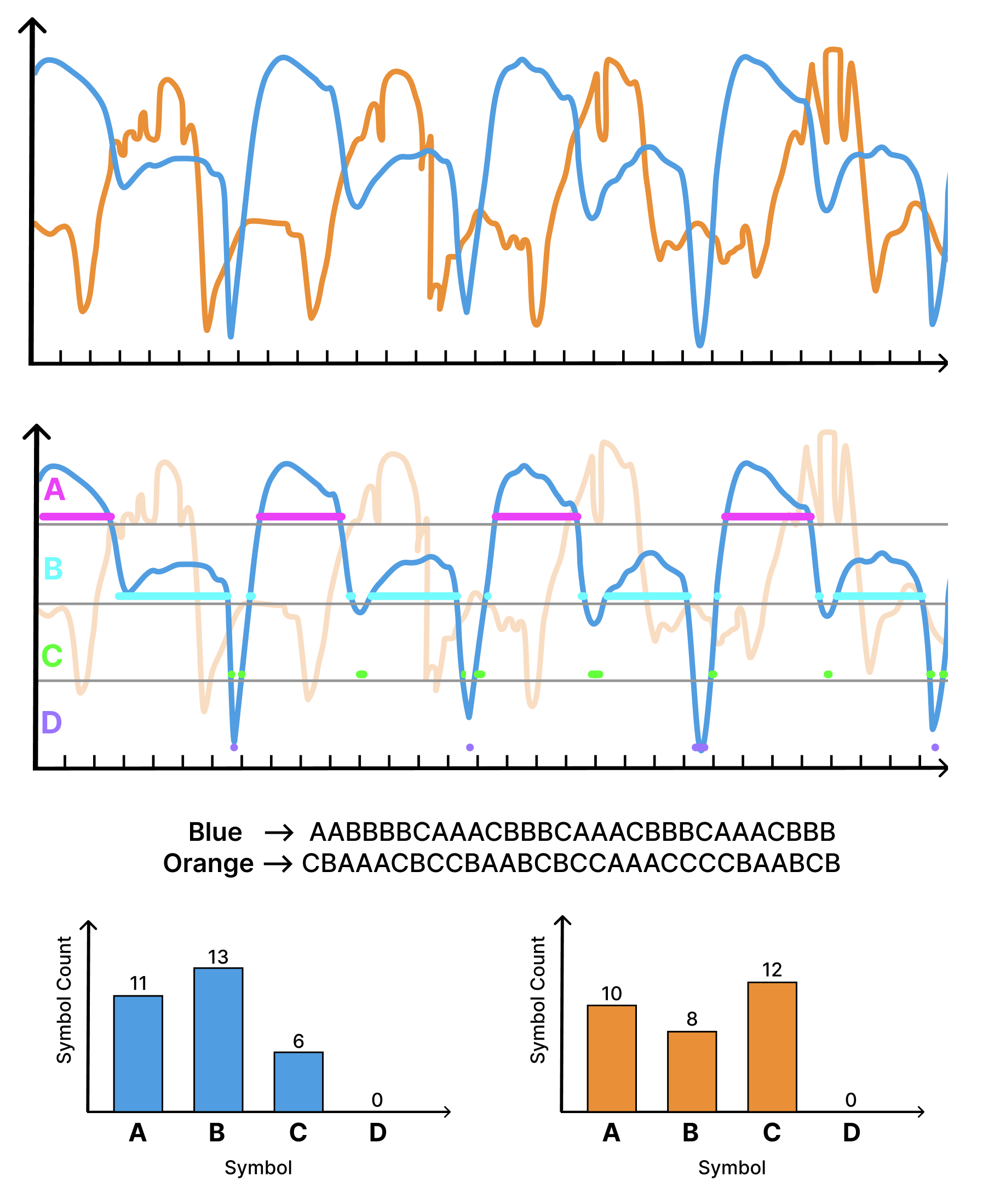}
    \caption{Sample Transformation to Symbolic Representation}
    \label{fig:one}
   
    \vspace{-4mm}
\end{figure}

Inspired from the bag-of-pattern style representations \cite{senin2013sax}, which is known to be insensitive to data warping, location shifts, and noise within time series data, we aim to help deep learning models acquire a representation resistant to such shifts. Our motivation is demonstrated in Figure \ref{fig:one}. Two activity time series\cite{reiss2012introducing} representing the same activity in two different people, shown in blue and orange respectively. Visually, these examples seem dissimilar due to data shifting, warping, and extra noise. But converting the samples into symbolic representations (Fig. 1.bottom) reveals their similarity. Driven from this observation, we propose a symbol-temporal consistency based SSL framework to address these challenges. The framework aims to encourage the representation consistency between embeddings generated from the symbolic representations and original temporal representations, forcing the model learn a warping-invariant representation. 

\section{Related Work}
\vspace{-2mm}

Recently, there has been a growing trend in designing self-supervised learning frameworks\cite{chen2020simple,grill2020bootstrap} for time series data ~\cite{yue2022ts2vec,franceschi2019unsupervised,tonekaboni2021unsupervised,eldele2021time}. Many of these models are based on the unsupervised contrastive learning framework SimCLR ~\cite{chen2020simple}. Yue et al. \cite{yue2022ts2vec} introduced a self-supervised learning framework named ts2vec, which employs time-stamp-wise and instance-wise contrastive co-learning to train the model. Additionally, Jiang et al. focused on enhancing SSL performance by utilizing both time and frequency domains \cite{jiang2020selfsupervised}. Zhang et al. \cite{zhang2022self} introduced a SSL framework named Time-Frequency Consistency (TFC) to perform a time-frequency domain independ self-supervised learning. Furthermore, other works focus on designing augmentation techniques. The existing literature explores various augmentation strategies such as jittering, scaling, masking and time-shifts to augment time-based embeddings \cite{franceschi2019unsupervised,tonekaboni2021unsupervised,eldele2021time,dong2023simmtm}. 

However, there has not been any SSL research that incorporates symbolic information. In the realm of time series data mining, symbolic representation techniques have traditionally been employed for tasks such as motif discovery\cite{gao2017efficient}, classification\cite{senin2013sax,gao2019hime}, and clustering\cite{megalooikonomou2005multiresolution} but they have yet to be utilized in SSL. This paper bridges the gap between symbolic and temporal domains, drawing inspiration from the symbolic representation counting paradigm to enrich contrastive learning\cite{abs-2005-13249} in time series data.

\section{Methodology}

\subsection{Problem Description}

Given a pre-training dataset $\mathcal{D}_{pret} = \{x_i \mid i = 1, \ldots, N\}$ which consists of $N$ unlabeled time series samples $x_i \in R^{K \times L}$. Self-supervised learning aims to leverage $\mathcal{D}_{pret}$ to train a model $f:x \rightarrow z$ such that by fine-tuning model parameters on a dataset $D_{tune}=\{x'_i \mid i = 1, \ldots, N'\}$ where $x'_i \in R^{K \times L}$, the fine-tuned model can adapt to various downstream tasks. Specifically, in the task of time series classification, we assume the labels of the data in $\mathcal{D}_{tune}$ are unknown and denoted as $y_{i} \in \{1, \ldots, C\}$.  
\vspace{-3mm}
\subsection{Overall Framework}
\vspace{-3mm}
\begin{figure}[!h]
    \centering
   \includegraphics[scale=0.25]{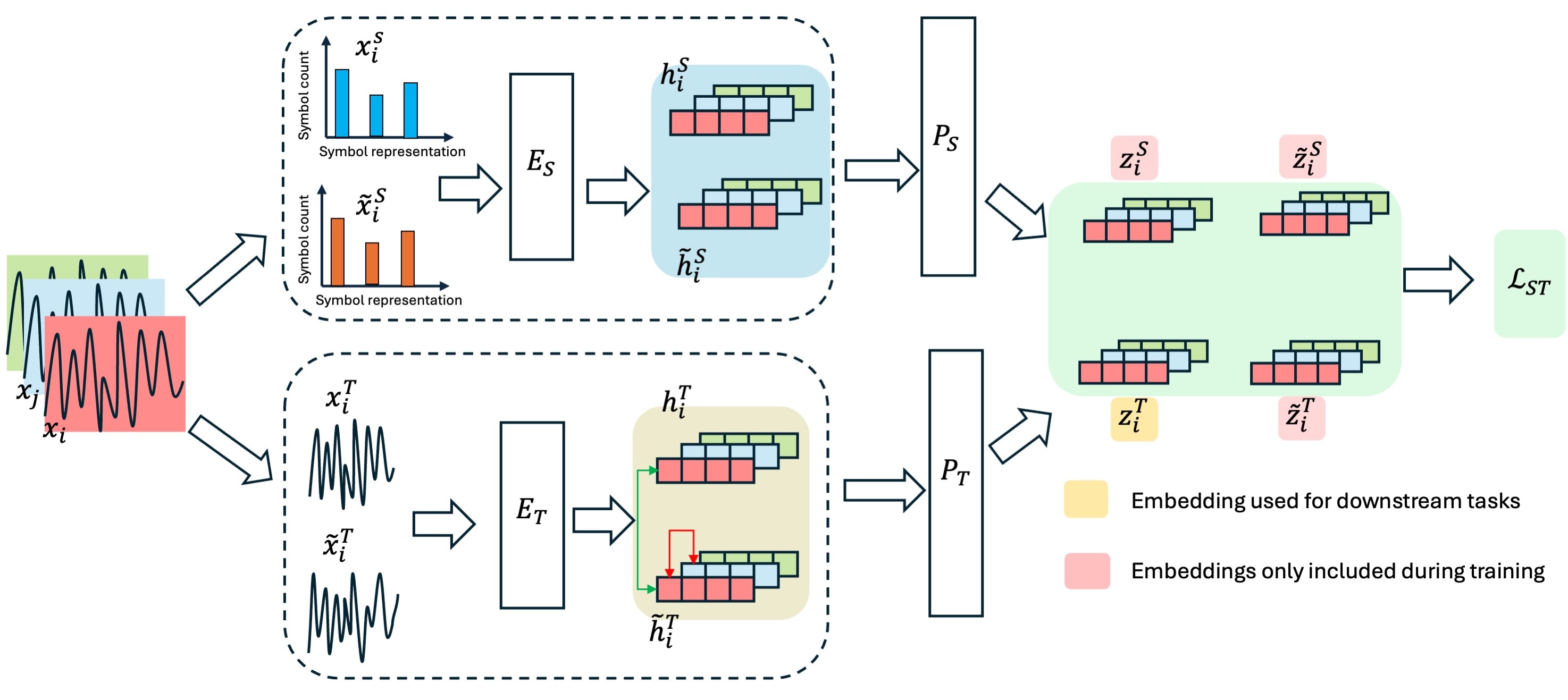}
    \caption{Overall Model Training Framework}
    \label{fig:enter-label}
\end{figure}
\vspace{-3mm}

The overall framework of our proposed method is shown in Figure \ref{fig:enter-label} and can be described by five steps: pre-processing, augmentation, encoding, self-contrastive learning, and symbol-temporal consistency learning. Given an input time series data, the framework first performs pre-processing to produce two copies of data representations: the bag-of-symbols representation, and its raw temporal representation. These representations are then augmented using symbol modification and random noise respectively. Next, each symbolic representation is+ passed through a symbolic encoder and each temporal representation is passed through a time encoder to obtain the embedding series representations. These embeddings are then used to perform self-contrastive learning on the symbolic and temporal embeddings independently to enhance robustness. Finally, embeddings from different data representations are aligned via symbol-temporal consistency loss. Through the above steps, we're able to guide the model to learn high-level features robust to both symbolic and temporal perturbations.

\subsection{Bag-of-Symbol Representation $x^{S}$:}

In addition to its raw temporal representation $x^T$, $x^T$ is transformed into a symbolic representation $x^S$ to better learn warping and shift invariance. For each element in $x^T$, we employ a discretization function that transforms real values into symbols. In this paper, the discretization function is designed to produce $n$ symbols, with $n-2$ of those symbols evenly spaced within the range of $\pm3\sigma$ (standard deviations) of the data with equal-width gap (i.e. $c_i=-3\sigma + \frac{6\sigma}{N-2}i$). By placing the remaining two symbols beyond this range, one above and one below, we account for all possible ranges, allowing for more accurate depictions on patterns in our data.

Specifically, given a time series data $x_{i,t,k}$ where $t$ represents the sample index, $k$ represents $k_{th}$ dimension, and cut lines $c_{j}$ where $j$ represents the symbol index, the discretization can be represented as follows:

\vspace{-3mm}
\begin{equation}
s_{i,t,k} =  i \quad c_i \geq x_{i,t,k} > c_{i+1} 
\end{equation}

We aggregate symbol counts within the discretized time series, ignoring order to simplify the representation and improve robustness to distribution shifts. The count is computed as:

\vspace{-3mm}
\begin{equation}
s_{i,j} = 
\sum_{t=1}^{L} \mathbb{I}\{ \text{symbol\_series}_{itk} = j \}
\end{equation}

\vspace{-2mm}
where $s_{ij}$ denotes the count of symbol $i$ in sample $j$, and $\mathbb{I}$ is the indicator function. By focusing on symbol frequency rather than temporal order, this approach reduces sensitivity to noise and suits tasks where timing is less critical. Thus, symbolic representations enhance the model’s capacity to capture key shifts and warping invariant behavior in time series data.

Lastly, to ensure a smooth training process, the scale of the count can be challenging to feed into deep learning models. Thus, $x_i^S$ is obtained by performing z-score normalization on top of $s_i$:

\vspace{-2mm}
\begin{equation}
    x_i^S = \frac{s_i - \mathbb{E}\{s_i\}}{\sigma_{i}^2}
\end{equation}
where $\sigma_{i}$ denotes the standard deviation of $s_i$.

\vspace{-2mm}
\subsection{Time-wise and Symbolic Representation Augmentation}

Following the classical SSL learning framework \cite{chen2020simple, zhang2022self}, $x_{i}^{s}$ and $x_{i}^{T}$ are augmented to $\tilde{x}_i^S$ and $\tilde{x}_{i}^T$ respectively. $\tilde{x}_{i}^{s}$ is produced through symbolic modifications (insertion or deletion) to $x_{i}^{s}$, and $\tilde{x}_{i}^{T}$ is produced by injecting noise into $x_{i}^T$.

\vspace{-2mm}
\subsection{Data Encoding }
\vspace{-1mm}

Next, we describe the proposed model which is trained based on the consistency loss structure introduced by Zhang et al. \cite{zhang2022self} to encourage the model to project time series with similar temporal or symbolic histogram characteristics into similar embedding representations. Specifically, $x_{i}^{T}$, $\tilde{x}_{i}^{T}$ pass through the time encoder to the latent embedding space $h_i^{T}$ and $\tilde{h}_{i}^{T}$, where ${h}_{i}^{T} = E_T(x_{i}^{T})$ and $\tilde{h}_{i}^{T} = E_T(\tilde{x}_{i}^{T})$. 
Similarly, the symbolic component $x_{i}^{S}$ and its augmentation sample $\tilde{x}_i^{S}$ are fed into the symbolic encoder $E_{S}$, mapping the samples to symbolic embeddings ${h}_{i}^{S} = E_S({x}_{i}^{S})$ and $\tilde{h}_{i}^{S} = E_S(\tilde{x}_{i}^{S})$. Next, we project all the embeddings into a unified latent space via the following functions: ${z}_{i}^{S} = P_S({h}_{i}^{S})$, $\tilde{z}_{i}^{S} = P_S(\tilde{h}_{i}^{S})$, ${z}_{i}^{T} = P_T({h}_{i}^{T})$, and $\tilde{z}_{i}^{T} = P_T(\tilde{h}_{i}^{T})$.

The creation of this joint time-symbolic space allows our model to unite time series and symbolic representations, allowing for improved generalization across diverse datasets. In the testing phrase, only $z_{i}^{T}$ will be treated as the model output to perform fine-tuning tasks.

\subsection{Self-contrastive Learning}

Inspired by prior work\cite{zhang2022self}, we propose a consistency contrastive loss with three components: a time-based loss using time-domain embeddings, a symbolic-based loss using bag-of-symbol embeddings, and a consistency loss aligning both. Positive pairs (original and augmented samples) are encouraged to be close in the embedding space, while negative pairs (original and other samples) are pushed apart. This encourages the model to capture underlying patterns from both time and symbolic views. Specifically, the time-based loss is described as:
\vspace{-3mm}
\begin{equation}
    L_{T,i} = -\log \frac{\exp(\text{sim}(h_{i}^{T}, \hat{h}_{i}^{T})/\tau)}{\sum_{j \in \mathcal{D}_{pret}} \mathbb{I}_{i \neq j} \exp(\text{sim}(h_{i}^{T}, G_T(x_j))/\tau)}
\end{equation}

where ${sim}(a,b) = {a^{T} b} / {\|a\| \|b\|}$ is the cosine similarity, $\mathbb{I}_{i \neq j}$ is an indicator function that equals 0 when $i = j$ and 1 otherwise, and $\tau$ is a temporal parameter used to adjust the scale. 

The symbol-based loss aims to ensure that the symbolic encoder $E_{S}$ generates embeddings that are invariant to symbolic perturbations by using positive and negative pairs to guide the model in learning symbolic representations. 
\vspace{-1mm}
\begin{equation}
    L_{S,i} = -\log \frac{\exp(\text{sim}(h_{i}^{S}, \tilde{h}_{i}^{S})/\tau)}{\sum_{j \in \mathcal{D}_{pret}} \mathbb{I}_{i \neq j} \exp(\text{sim}(h_{i}^{S}, E_S(x_j))/\tau)}
\end{equation}

\subsection{Symbol-Temporal Consistency Learning}

Lastly, the time-symbolic loss enforces consistency between the time-based $z_i^T, \Tilde{z}_i^T$ and symbolic-based embeddings $z_i^S, \Tilde{z}_i^S$. We first compute all the across domain contrastive loss:
\begin{equation}
    L_{a,b} = -\log \frac{\exp(\text{sim}(a,b)/\tau)}{\sum_{j \in \mathcal{D}_{pret}} \mathbb{I}_{i \neq j} \exp(\text{sim}(a,b))/\tau}
\end{equation}

and the consistency loss is defined as:
\vspace{-1mm}
\begin{equation}
L_{TS,i} = \sum_{a \in \{z_{i}^T,\tilde{z}_{i}^T\}}\sum_{b \in \{z_{i}^S,\tilde{z}_{i}^S\}} (L_{t,s} - L_{a,b} + \delta)    
\end{equation}
where $\delta$ denotes the hyper-parameter.  Finally, the overall loss function is:
\vspace{-2mm}
\begin{equation}
    L = \sum_{x_i \in \mathcal{D}_{pret}}L_{T,i} + L_{S,i} + \lambda L_{TS,i}
\end{equation}
\vspace{-1mm}
where $\lambda$ allows us to control the importance of $L_{TS,i}$.

Jointly optimizing these losses enables the model to learn embeddings that are robust to noise, adaptable to perturbations, and rich in temporal and symbolic patterns, boosting downstream performance and generalizability.

\vspace{-1mm}
\section{Experimentation}

We evaluate our implementation through comparison experiments conducted on a T4 Nvidia GPU. Using the Adam optimizer (learning rate: $5e - 4$) and a ReduceLROnPlateau scheduler, we base our experiments on the PAMAP2 dataset. Our discretization function produces 64 symbols using 62 cutlines. For MLP and Linear Regression testing, we use only the $z_t$ embeddings, as shown in Figure 2, to obtain accuracy.

\begin{table}[h]
\centering
\resizebox{0.8\columnwidth}{!}{%
\begin{tabular}{cc|cccc}
Source        & Target        & Proposed                                           & TFC                                                           & MLP                                                  & LR                                                   \\ \hline
2             & 1             & \textbf{0.909}                                                & \uline{0.825}                                                   & 0.662                                                & 0.352                                                \\
5             & 1             & \uline{0.740}                                                 & \textbf{0.908}                                                & 0.433                                                 & 0.389                                              \\
6             & 1             & \uline{0.932}                                                   & 0.878                                                         & \textbf{0.953}                                       & 0.636                                                \\
8             & 1             & \textbf{0.910}                                                & \uline{0.863}                                                   & 0.819                                                & 0.562                                                \\ \hline
1             & 2             & \textbf{0.896}                                                & \uline{0.864}                                                   & 0.605                                                & 0.127                                                \\
5             & 2             & \uline{0.888}                                                 & \textbf{0.918}                                                 & 0.544                                                  & 0.137                                                \\
6             & 2             & \textbf{0.915}                                                & \uline{0.878}                                                   & 0.137                                                & 0.137                                                \\
8             & 2             & \textbf{0.954}                                                & \uline{0.934}                                                   & 0.572                                                & 0.122                                                \\ \hline
1             & 5             & \uline{0.853}                                                   & \textbf{0.922}                                                & 0.611                                                & 0.399                                                \\
2             & 5             & \textbf{0.920}                                                & \uline{0.888}                                                   & 0.654                                                & 0.347                                                \\
6             & 5             & \textbf{0.955}                                                & \uline{0.883}                                                   & 0.639                                                & 0.295                                                \\
8             & 5             & \uline{0.908}                                                   & \textbf{0.935}                                                & 0.593                                                & 0.336                                                \\ \hline
1             & 6             & \textbf{0.922}                                                & \uline{0.874}                                                   & 0.576                                                & 0.297                                                \\
2             & 6             & \textbf{0.886}                                                & \uline{0.855}                                                  & 0.450                                                & 0.328                                               \\
5             & 6             & \textbf{0.959}                                                & \uline{0.907}                                                   & 0.825                                                & 0.639                                                \\
8             & 6             & \textbf{0.938}                                                & \uline{0.929}                                                   & 0.827                                                & 0.583                                                \\ \hline
1             & 8             & \textbf{0.905}                                                & \uline{0.825}                                                   & 0.662                                                & 0.352                                                \\
2             & 8             & \uline{0.819}                                                & \textbf{0.908}                                                  & 0.433                                                & 0.389                                                \\
5             & 8             & \uline{0.899}                                                & 0.878                                                           & \textbf{0.953}                                          & 0.636                                                \\
6             & 8             & \uline{0.929}                                                & 0.885                                                           & \textbf{0.930}                                        & 0.562                                                \\ \hline
\multicolumn{2}{c|}{Average:} & \textbf{0.901}                                                & \uline{0.888}                                                   & 0.644                                                & 0.381                                               
\end{tabular}%
}
\caption{Baseline comparisons. Source and Target correspond to Subject ID, where the model is trained on the Source data and tested on the Target Data.}
\end{table}

\begin{table}[h]
\centering
\resizebox{0.5\columnwidth}{!}{%
\begin{tabular}{cc|cc}
Source        & Target        & $z_t$                                                  &  $z_t$ + $z_s$                                                            \\ \hline
2             & 1             & \textbf{0.926}                                                & 0.853                                                         \\
5             & 1             & \textbf{0.948}                                                & 0.920                                                         \\
6             & 1             & 0.923                                                         & \textbf{0.955}                                                \\
8             & 1             & \textbf{0.912}                                                & 0.908                                                         \\ \hline
1             & 2             & 0.872                                                         & \textbf{0.896}                                                \\
5             & 2             & 0.833                                                         & \textbf{0.888}                                                \\
6             & 2             & 0.42                                                          & \textbf{0.915}                                                \\
8             & 2             & 0.882                                                         & \textbf{0.954}                                                \\ \hline
1             & 5             & 0.887                                                         & \textbf{0.922}                                                \\
2             & 5             & \textbf{0.888}                                                & 0.886                                                         \\
6             & 5             & \textbf{0.966}                                                & 0.959                                                         \\
8             & 5             & 0.925                                                         & \textbf{0.938}                                                \\ \hline
1             & 6             & 0.808                                                         & \textbf{0.905}                                                \\
2             & 6             & 0.794                                                         & \textbf{0.847}                                                \\
5             & 6             & \textbf{0.881}                                                & 0.874                                                         \\
8             & 6             & 0.839                                                         & \textbf{0.905}                                                \\ \hline
1             & 8             & 0.905                                                         & \textbf{0.909}                                                \\
2             & 8             & \textbf{0.819}                                                & 0.740                                                         \\
5             & 8             & 0.899                                                         & \textbf{0.932}                                                \\
6             & 8             & \textbf{0.929}                                                & 0.905                                                         \\ \hline
\multicolumn{2}{c|}{Average} & \textbf{0.901}                                                & 0.863                                                        
\end{tabular}%
}
\caption{$z_t$+$z_s$ combined embedddings vs $z_t$ embeddings only}
\vspace{-5mm}

\end{table}

\noindent \textbf{Dataset characteristics}:
We use the PAMAP2 dataset \cite{reiss2012introducing}, a Human Activity Recognition dataset, to evaluate the performance of our model. The PAMAP2 dataset comprises recordings from 9 subjects, capturing various activities. In our experiments, we focus on a  classification task specifically designed for classifying standing, walking, and running. To ensure meaningful evaluations, we only included subjects with at least 90 seconds of recorded data in all three activities, resulting in the selection of 5 our of 9 candidates (subjects 1, 2, 5, 6, 8). The experiment is conducted using pairwise classification tasks, where the model is trained on data from candidate $i$ and tested on candidate $j$. Following the evaluation protocol in \cite{yue2022ts2vec}, after training the model using the SSL framework, we use the embeddings to train a logistic regression model for performance comparison. We compared our proposed work with 3 baselines: \textbf{TFC}\cite{zhang2022self}: Time-Frequency Consistency based SSL, and two models that training from scrap,  Multi-layer perceptron (\textbf{MLP}) and Logistic Regression (\textbf{LR}).
\vspace{-2mm}
\subsection{Result}
\noindent \textbf{Comparison Result}: The results, compared with all baselines, are shown in Table I . The best-performing implementation is bolded in Table I , and the second best is underlined. From the table, our implementation consistently performs well against the TFC baseline method. It is notably superior when the target is subject 2, supporting our claim our implementation is useful for addressing the distribution offset often characterized within genders. The result demonstrates that the proposed approach can outperform the existing self-supervised learning framework.

\noindent \textbf{Ablation Test}:
Next, we conduct an ablation test to evaluate the effectiveness of only using $z_s$ in the proposed framework. The results of using a combination of $z_s$ and $z_t$, the strategy used in the TFC framework, along with our performance, are shown in Table II . Throughout experimentation, we do note that when omitting embeddings $z_s$, we observe an improved performance when compared to combined $z_t$ and $z_s$ embeddings. This leads us to believe that by omitting Symbolic $z_s$ embeddings during testing, we can achieve better performance. 
\vspace{-4mm}
\section{Conclusion}

    In this paper, we proposed a Self-Supervised Learning framework called Symbol-Time Consistency(STC), which integrates the bag-of-symbol representation and proves robustness against warping, location shifting, and noise within data. We demonstrate that the proposed work can enhance the generalization of deep learning models for time series, specifically in the task of activity recognition. Through extensive experiments, we demonstrate the effectiveness of our framework, particularly in handling distribution shifts arising from characteristics of unseen users. 
   
\section{Acknowledgements}

This work is supported by Google CASHI IRP and National Science Foundation (NSF) under Grant IIS-2348480. The computational resource is supported by Google Cloud Platform (GCP) credit. 

\bibliography{main}
\bibliographystyle{plain}
\end{document}